\pdfoutput=1

\documentclass[11pt]{article}

\usepackage[]{acl}

\usepackage{times}
\usepackage{latexsym}

\usepackage[T1]{fontenc}

\usepackage[utf8]{inputenc}

\usepackage{microtype}

\usepackage{inconsolata}

\usepackage{graphicx}

\usepackage[frozencache,cachedir=.]{minted}

\usepackage{listings}
\title{IMGTB: A Framework for Machine-Generated Text Detection Benchmarking}

\author{Michal Spiegel$^{1,2}$ \and Dominik Macko$^1$\\
  $^{1}$ Kempelen Institute of Intelligent Technologies \\
  $^{2}$ Faculty of Informatics, Masaryk University\\
 \texttt{michal.spiegel@intern.kinit.sk}, \texttt{dominik.macko@kinit.sk} \\}

\begin{document}
\maketitle
\begin{abstract}
In the era of large language models generating high quality texts, it is a necessity to develop methods for detection of machine-generated text to avoid harmful use or simply due to annotation purposes. It is, however, also important to properly evaluate and compare such developed methods. Recently, a few benchmarks have been proposed for this purpose; however, integration of newest detection methods is rather challenging, since new methods appear each month and provide slightly different evaluation pipelines.
In this paper, we present the IMGTB framework, which simplifies the benchmarking of machine-generated text detection methods by easy integration of custom (new) methods and evaluation datasets. Its configurability and flexibility makes research and development of new detection methods easier, especially their comparison to the existing state-of-the-art detectors. The default set of analyses, metrics and visualizations offered by the tool follows the established practices of machine-generated text detection benchmarking found in state-of-the-art literature.
\end{abstract}

\section{Introduction}

Due to indistinguishability between human-written texts and high-quality texts generated by modern large language models (LLMs) \citep{sadasivan2023aigenerated}, the machine-generated text detection (MGTD) belongs to the key challenges identified by \citep{kaddour2023challenges}. MGTD methods are needed in many areas, such as prevention of disinformation spreading, plagiarism, impersonation and identity theft, automated scams and frauds, or even prevention of unintentional inclusion of lesser quality generated texts in future models’ training data \citep{kaddour2023challenges, weidinger2021ethical, zellers2019neuralfakenews, wahle-etal-2022-large, vykopal2023disinformation}.

\begin{figure}[!t]
    \centering
    \includegraphics[width=\linewidth,trim=6cm 7cm 6cm 7cm, clip]{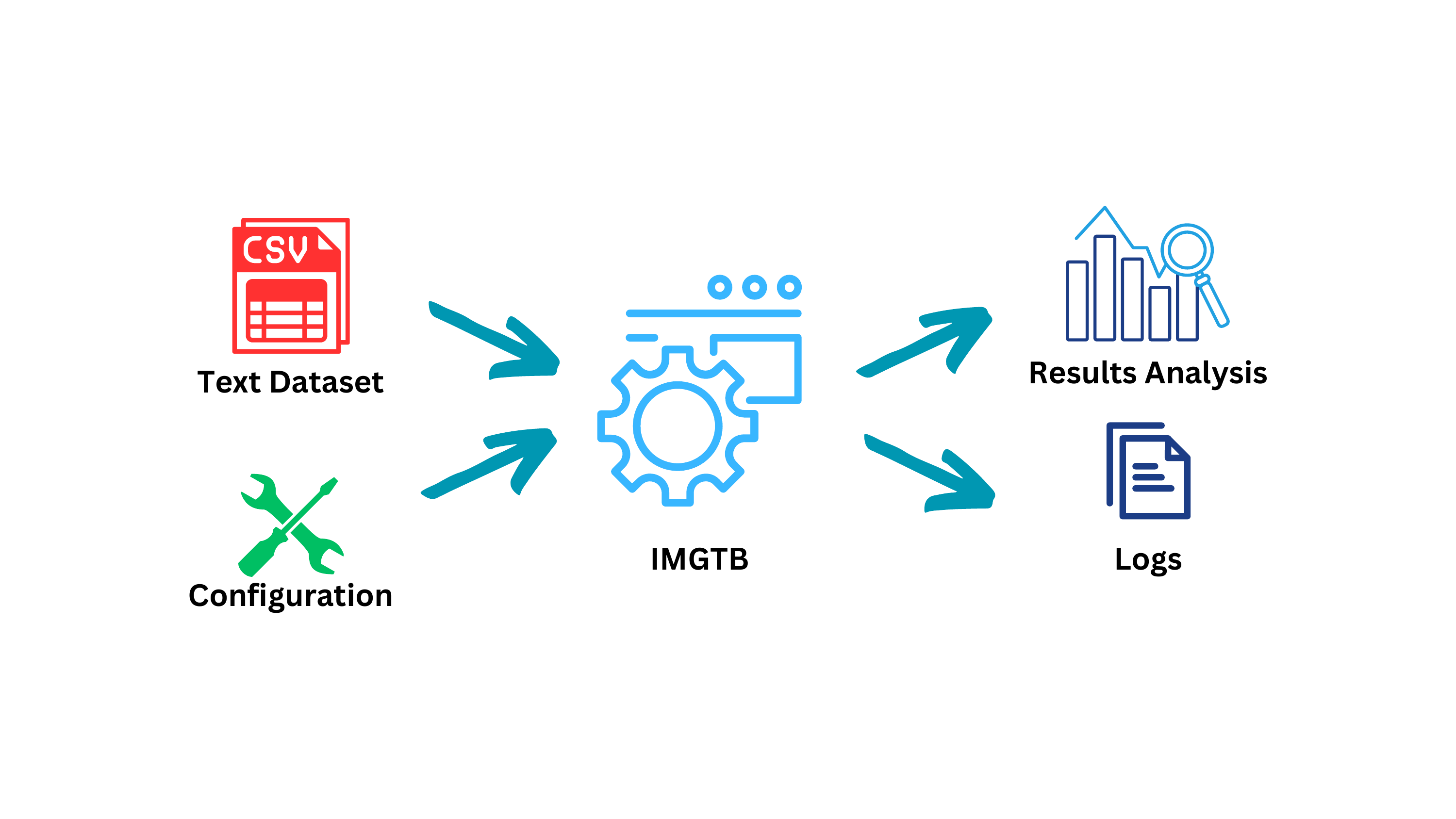}
    \caption{IMGTB framework exemplar usage overview.}
    \label{fig:teaser}
    \vspace{-3mm}
\end{figure}

Regardless of the area of application, we are witnessing a race of new MGTD methods competing with new \textit{generation} methods and appearing monthly.
This presents a challenge to efficiently evaluate and benchmark the new methods. The problem is twofold, missing the uniform implementation of the methods and standardized evaluation. Even when source codes for experiment replication are released, they are usually too specific and not flexible for reuse. Moreover, across application areas, domains, text lengths, or topics, the performance of different MGTD methods varies. Therefore, a flexible way of comparison over various datasets (even custom ones) is currently missing. These problems are usually addressed by common benchmarking frameworks.

There is a lack of flexibility, configurability, and extensibility in the current MGTD benchmarking frameworks; therefore, we have focused on refining the most recent one, MGTBench \citep{he2023mgtbench}, by integrating missing features. The key contributions of the proposed extended framework IMGTB\footnote{\url{https://github.com/michalspiegel/IMGTB}} are as follows:
\begin{itemize}
    \setlength{\parskip}{0pt}
    \item simplified \textit{implementation} of new MGTD methods (by abstract classes and templates),
    \item more flexible usage of custom evaluation \textit{datasets} (multi-format support),
    \item increased \textit{configurability} of the benchmark settings (e.g., classifier selection),
    \item benchmark results \textit{analysis} (configurability, automated charts generation).
\end{itemize}

The paper is organized as follows. In the next section, the related works representing the state-of-the-art (SOTA) are analyzed. In Section~\ref{sec:framework}, the proposed extended framework for MGTD benchmarking is introduced. Section~\ref{sec:case_study} describes a case study, providing exemplar real usage of the framework. In Section~\ref{sec:extension}, the possibilities for further enhancements of the framework are identified. The last section concludes the results.

\section{Related Works}
\label{sec:related_works}

Due to increasing quality of texts generated by modern LLMs, the research around detection of machine-generated text increased its importance. However, a common way to properly compare several detection methods was missing, mainly due to missing publicly available datasets. Few years ago, MGTD researchers mostly used data generated by a single LLM, such as GPT-2\footnote{\url{https://github.com/openai/gpt-2-output-dataset}} or Grover \citep{zellers2019neuralfakenews}, results on which could not be properly generalized. Later on, larger-scale multi LLM benchmarks for MGTD task have been proposed, such as TuringBench \citep{uchendu-etal-2021-turingbench-benchmark}, DeepfakeTextDetect \citep{li2023deepfake}, M4 \citep{wang2023m4}, or MULTITuDE \citep{macko2023multitude}. As a result, MGTD methods can now be evaluated on such benchmark datasets and compared to each other. However, these datasets do not share common class labels, structure, or form, what makes the evaluation on multiple of them complicated and unnecessarily prolongs the research.

The other issue significantly prolonging the research is a missing unified implementation of existing MGTD methods. It leaves on the researchers a burden to either reuse the published source codes of individual methods (if there is some), which are different among each other and require customization, or implement them completely into their evaluation framework to be evaluated in a unified way with their newly proposed MGTD method. Some of the proposed MGTD methods, such as DetectGPT \citep{mitchell2023detectgpt}, released the full source code including implementation of other existing SOTA methods, enabling complete replication of experiments and providing a good basis to build upon. The result is a faster advancement by extension of the original method, in the form of DetectLLM \citep{su2023detectllm} or Fast-DetectGPT \citep{bao2023fast}, proving the benefits of full replication possibilities.

However, these methods focused on zero-shot statistical-based detection of machine-generated text, comparing various statistical metrics to distinguish between human-written and machine-generated samples, not providing the classification prediction. Thus, the implementations do not allow easy comparison with supervised high-performing pretrained LLMs finetuned for MGTD task, such as the popular OpenAI detector \citep{solaiman2019release}. The proposed MGTBench framework \citep{he2023mgtbench} attempted to solve the problem, by implementation of these methods in a common framework. Using a dedicated classifier trained individually for metric-based statistical MGTD methods, it provides a class prediction, enabling a direct comparison to LLMs-based MGTD classifiers. MGTBench has already accelerated MGTD research, such as \citep{wu2023mfd} or \citep{macko2023multitude}. However, it provides a quite complicated way to use custom datasets or to integrate new MGTD methods.

\section{IMGTB -- Integrated MGTD Benchmark Framework}
\label{sec:framework}
In this section, we introduce the central design principles, IMGTB was built with, as well as its architecture and the functionality of the main components.
We use a term \textit{experiment} to denote a single run of the specified detection method on data from the specified dataset.

\begin{figure*}
    \centering
    \includegraphics[width=\textwidth]{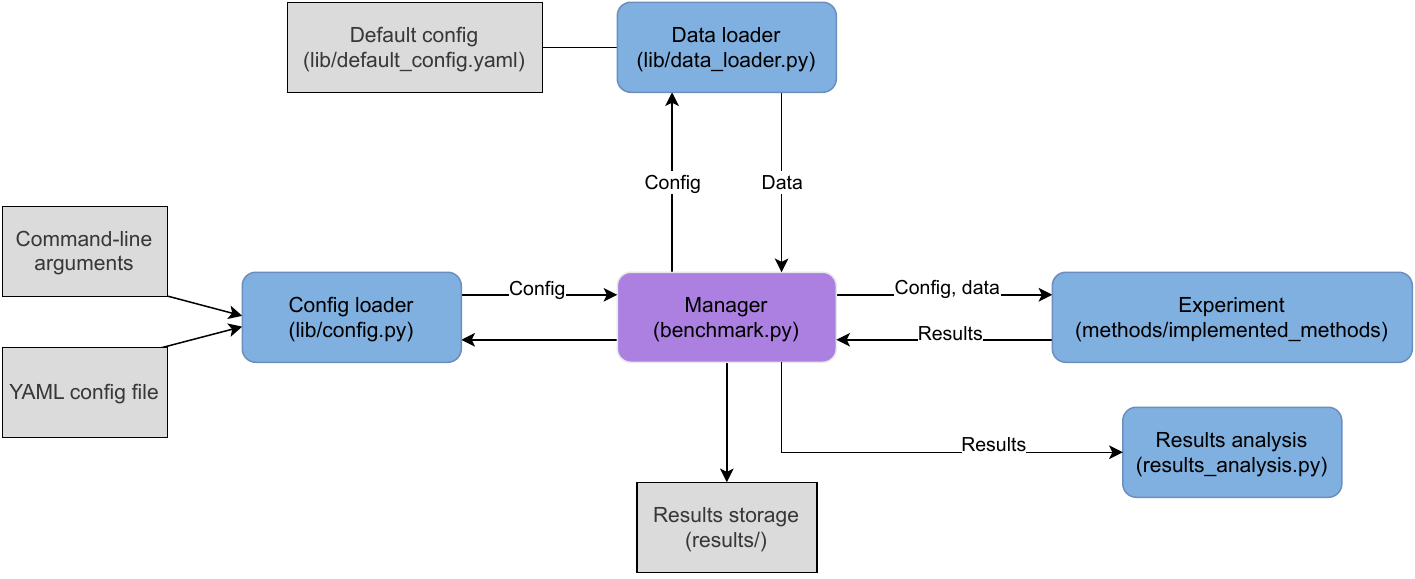}
    \caption{IMGTB architecture overview.}
    \label{fig:archi}
\end{figure*}

\subsection{Design Principles}
The IMGTB framework was designed with several main principles in mind. We consider them important to mention because they encompass what was missing in other similar works and why this tool was developed in the first place.

\paragraph{[P1: Modularity]}
All the subtasks and responsibilities, such as configuration parsing, data loading, and runnning experiments, were divided and assigned to their respective modules that only communicate between themselves through a very general interface. Such a decreased inter-module coupling makes the framework very robust and resistant to changes and easy to update, which is useful in order to utilize all the technologies that are yet to be discovered.

\paragraph{[P2: Ease of use]}
The issue and a main blocker when testing and experimenting with new MGTD methods and new datasets seems to be the need to manually set up and integrate a new method, which often does not work out-of-the-box, to manually parse each dataset and then write your own analysis tools. This framework was designed to mitigate this issue. Simple experiments can be running in seconds just using the terminal via command-line arguments or, for more complex experiments, using a YAML configuration file. Any dataset or detector can be easily accessed from the Hugging Face Hub without the need to manually download it. The framework also includes many parsing utility functions that enable to load and parse almost any dataset without any need to provide a custom code. Additionally, with built-in analysis tools, it is possible to have basic analysis done right after the experiment has finished.

\paragraph{[P3: Customizability]}
The structure of input data can vary significantly, detection methods often need different resources, and although we do try to provide utility functions to provide for most of them, it is not possible to cover all such possible cases. Therefore, we have put great emphasis on making the customization of our codebase and extending our functionalities as simple and straightforward as possible.

\subsection{Architecture Overview}
\figurename~\ref{fig:archi} overviews the main components of the framework architecture, further described in the following subsections.
\subsubsection{Manager}
The Manager, interconnecting all the other components, serves as the user interface. Its main task is to orchestrate the other components. It calls the data loader, forwards configurations, runs experiments and so on. 

\subsubsection{Configuration Parser}
Configuration Parser provides the functionality to specify configurations directly in the terminal via command-line arguments for quick experiment setup or via a YAML configuration file for more complex experiments. However, command-line arguments offer only a subset of the options the YAML configurations system offers. For convenience, user-specified configurations are always merged with a system default configurations (see \textit{lib/default\_config.yaml}). To add a new parameter to the configurations is as easy as adding it to your YAML configurations file, or to the system default (\textit{lib/default\_config.yaml}), no changes to the code itself are needed. 

\subsubsection{Data loader}
Data loader's main responsibility is to offer functionalities to parse as many different dataset formats and structures as possible. Currently, it is possible to specify column names, labels, a Hugging Face Hub dataset just by providing its identifier, use different subsets, splits, test on machine or human only text data, and much more. 
In the case that these predefined functionalities would not be sufficient, we try to make it as easy as possible to integrate custom parser functions.

\subsubsection{Experiment}
Experiment is an abstract class defining a single abstract method \textit{run(data, config)} that runs the experiment on the provided data and given configurations and returns results (ideally in the standardized format). In regards to detectors, the framework offers many already implemented (\textit{methods/implemented\_methods}), such as metric-based methods (e.g., Entropy by \citealp{10.5555/3053718.3053722}, or GLTR by \citealp{gehrmann2019gltr}) or perturbation-based methods (e.g., DetectGPT by \citealp{mitchell2023detectgpt}, or DetectLLM-NPR by \citealp{su2023detectllm}). To run a Sequence Classification Hugging Face Hub model, only its identifier needs to be specified in the methods configurations as a file path.
Although there are many MGTD methods already implemented in the framework, the true feature of this component is the possibility to quickly implement new custom experiments. By using some of the predefined experiment templates for metric-based or perturbation-based methods, it is possible to implement experiments in just a few lines of code. There is, however, still a possibility to implement a fully custom experiment by implementing the \textit{run()} method from scratch.

\subsubsection{Results Analysis}
Results analysis can be run either right after a benchmark run, can be specified in the configurations, or later by loading the results from a file. We implement several analysis methods ourselves, such as detection performance (Accuracy, Precision, Recall, F1-score), false positives/negatives, or run-time performance. But it is ensured for easy integration of new analysis methods.

\section{Case Study}
\label{sec:case_study}

To better illustrate the use of the framework in practice, in this section we showcase a few example use case scenarios. We look at:
\vspace{-2mm}
\begin{itemize}
    \setlength{\parskip}{0pt}
    \item[] \textbf{A.} How to quickly run and evaluate simple experiments using CLI
    \item[] \textbf{B.} How to run complex experiments using YAML configuration files
\end{itemize}
\vspace{-2mm}
For a more visual version of this demonstration, see this video\footnote{\url{https://www.youtube.com/watch?v=NlHIC4HDQrc}}.
For more detailed and runnable version, see this Jupyter notebook\footnote{\url{https://colab.research.google.com/drive/15C7kzpnDnx_zqwplCpc949xVJ4Bhdnjl?usp=sharing}}.

\subsection{Example Scenario A}
Let's assume we obtained a completely new never-before-seen dataset of texts generated by one of the latest SOTA large language models.
In a similar manner, we could also use existing datasets, even from completely unrelated domains, such as AI-powered text summarization, translation, question answering, or disinformation detection.

Out of curiosity, we'd like to see how the current SOTA detection methods roughly perform on this new data.

Starting from scratch, this would probably take a significant amount of effort to preprocess the data, find the source code of the detectors, integrate the detectors, evaluate and plot the results, as well as considerable knowledge about tools like pandas, numpy or transformers, not to mention the time spent browsing the documentation of said tools.

This all seems a little bit too much. But with our framework we could accomplish the same just by running one CLI command as follows:
\vspace{-1mm}
\begin{minted}[xleftmargin=0.1\columnwidth,fontsize=\small,breaklines]{bash}
python benchmark.py --dataset xzuyn/futurama-alpaca huggingfacehub machine_only output --methods roberta-base-openai-detector Hello-SimpleAI/chatgpt-detector-roberta andreas122001/roberta-mixed-detector
\end{minted}
\vspace{-1mm}
In the command, the option \textit{{-}{-}dataset} is used for specification of \textit{xzuyn/futurama-alpaca} dataset, available at HuggingFace (see the \textit{huggingfacehub} keyword), which contains only machine-generated texts (see the \textit{machine-only} keyword), and the data field/column to be used for texts being \textit{output}. For the full description of the \textit{dataset} parameters see the GitHub repository\footnote{\url{https://github.com/michalspiegel/IMGTB/tree/main\#dataset-parameters}}.
The option \textit{{-}{-}methods} is followed by identifiers of the methods to be evaluated and compared in the benchmark. If such identifiers are not found in the local implementations of the MGTD methods, the HuggingFace is used as a repository of the models.

\begin{figure}[!t]
    \centering
    \includegraphics[width=1\linewidth,trim=2.5cm 0cm 0cm 0cm, clip]{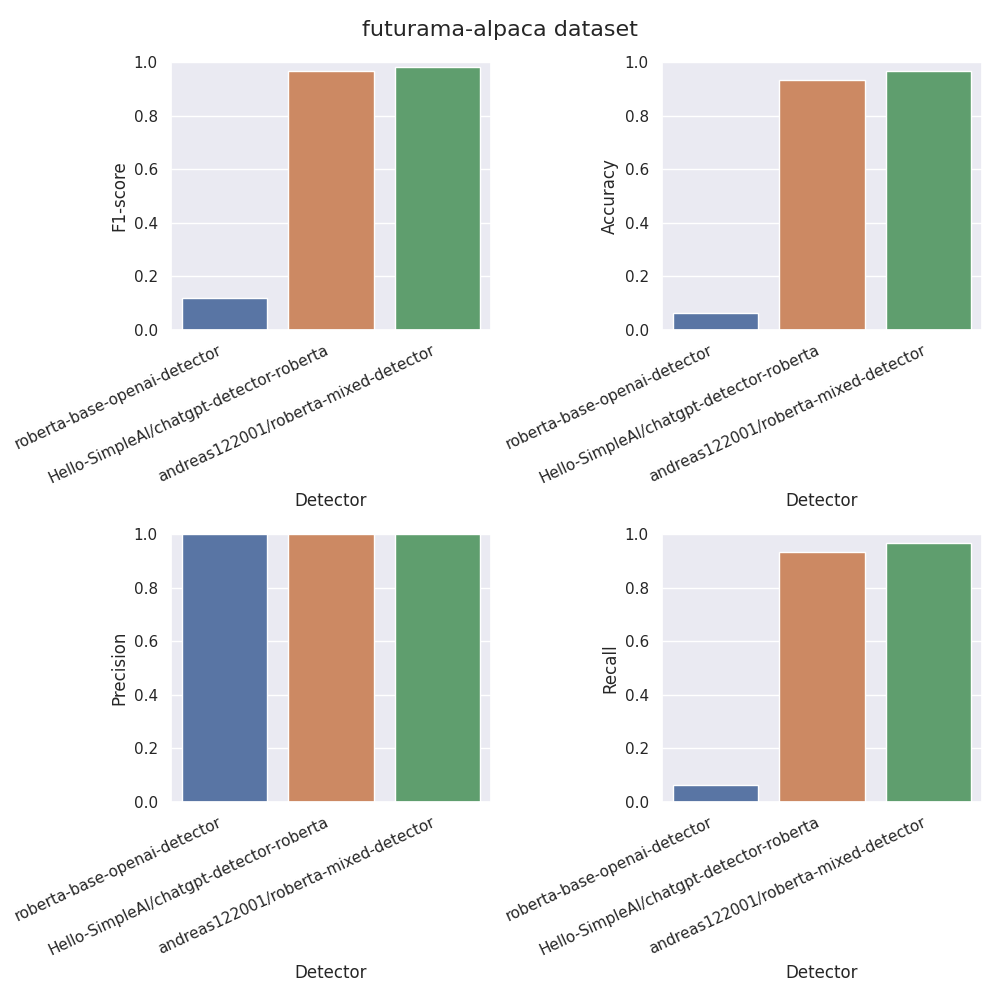}
    \caption{Automatically generated chart for detection-performance metrics analysis.}
    \label{fig:metrics}
\end{figure}

\begin{figure}[!t]
    \centering
    \includegraphics[width=1\linewidth,trim=3cm 0cm 0cm 0cm, clip]{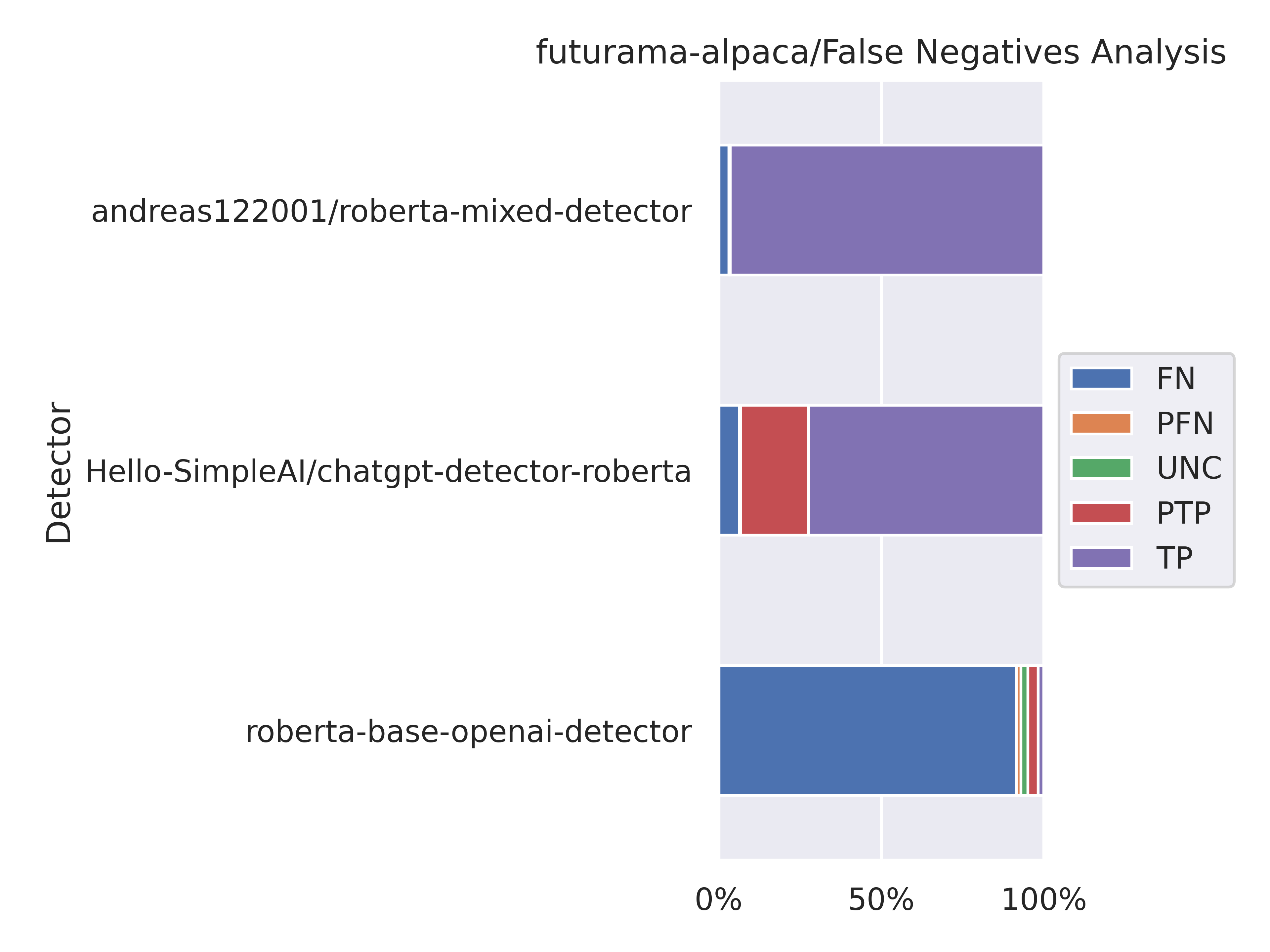}
    \caption{Automatically generated chart for false-negatives analysis. $FN$ represents false negatives, $PFN$ represents potential false negatives, $UNC$ represents uncertainty, $PTP$ represents potential true positives, and $TP$ represents true positives.}
    \label{fig:fn}
\end{figure}

When the benchmark run finishes, we are able to find all the results in the latest \textit{results/logs} log entry. It contains a JSON file storing all the benchmark results and the output plots (examples in \figurename~\ref{fig:metrics} and~\ref{fig:fn}) of the results analysis component. Using the provided plots, per-detection-method performance is easily comparable.

Regarding \figurename~\ref{fig:metrics}, only machine-class samples were included in the Scenario A dataset; therefore, the precision of all detectors is $1.0$ (i.e., no false positives) and the accuracy is the same as the recall. Based on \figurename~\ref{fig:fn}, the last detection method clearly has problems in identifying machine texts from the provided dataset, due to prevalence of false negatives (with a high certainty, based on machine-class probability score).

\subsection{Example Scenario B}
In this scenario, let's assume that we have developed and somehow integrated a new metric-based MGT detection method called \emph{MiracleMetric}.
To make a complex evaluation on multiple datasets, comparing with multiple different detection methods, and with different parameters, we can design a very compact and readable YAML configuration file.

Firstly, in \figurename~\ref{fig:yamldata} we specify the data to be used.
After that we can specify multiple methods (including our \emph{MiracleMetric}) with different parameters, models, etc. in \figurename~\ref{fig:yamlmethods}.

\begin{figure}[!t]
\centering
\begin{minted}[
    gobble=2,
    frame=single,
    fontsize=\small
  ]{yaml}
  data:
    global:
      filetype: auto
    list:
    - filepath: WxWx/ChatGPT-Detector-Bias
      filetype: huggingfacehub
      text_field: text
      label_field: kind
      human_label: Human-Written

    - filepath: yaful/DeepfakeTextDetect
      filetype: huggingfacehub
      train_split: test_ood_gpt
      test_split: test_ood_gpt_para
      human_label: "1"
\end{minted}
\vspace{-5mm}
\caption{Data configurations in YAML format.}
\label{fig:yamldata}
\end{figure}

\begin{figure}[!t]
\centering
\begin{minted}[
    gobble=2,
    frame=single,
    fontsize=\small
    ]{yaml}
  methods:
    global:
      base_model_name: gpt2-medium
      mask_filling_model_name: t5-large
      DEVICE: cuda
    list:
    - name: MiracleMetric
    - name: MiracleMetric
      clf_algo_for_threshold:
        name: MLPClassifier
        hidden_layer_sizes: [64, 32, 16]
    - name: LoglikelihoodMetric
    - name: LogRankMetric
    - name: EntropyMetric
    - name: DetectLLM_LLR
    - name: EntropyMetric
    - name: roberta-base-openai-detector
\end{minted}
\vspace{-5mm}
\caption{Methods configurations in YAML format.}
\label{fig:yamlmethods}
\end{figure}

With this done, the only step keeping us from the results is running the benchmark using these configurations:

\begin{minted}[xleftmargin=0.1\columnwidth,fontsize=\small,breaklines]{bash}
python benchmark.py --from_config=example_config.yaml
\end{minted}

This will output similar results to the previous example scenario. However, this time we might not be satisfied with the simple automatic analysis provided to us by the framework and we might want to do a more complex and custom-made analysis fitting to the specific needs of our benchmark run.
For a demonstration on this exact issue, see the provided Jupyter notebook with the full demo.

\section{Extension \& Enhancements Possibilities}
\label{sec:extension}

There are various limitations of the current version of the framework, which can be targeted to increase its usability even more. For example, automated dataset split selection is currently available from configuration file only; however, it is possible to extend such a support to CLI as well. The classifier of statistical metric-based or perturbation-based methods is currently trained from scratch in each experiment. It is possible to extend the framework in order to dump a trained classifier to be used in another experiment. This could bring a possibility to also use the statistical methods for test-only scenario (currently only the pretrained supervised classification models support this feature).

The \textit{bitsandbytes} library has already been utilized for $int8$ inference of perturbation models. It can be further used also for base models and utilize also even more GPU-efficient $int4$ inference. This could also boost the speed. Further speed-ups can be achieved by analyzing all experiments to be executed based on the configuration and schedule the experiments for the benchmark in order to eliminate redundant tasks (e.g., loading of the same base models for multiple methods, calculating the same metrics or generating the same perturbations for multiple methods).

There are even possibilities for significant extension of the framework beyond the current scope. For example, similarly to detection methods, authorship obfuscation methods (i.e., authorship hiding, evading detection) can be integrated into the framework to offer automated evaluation of adversarial robustness of the detection methods in the benchmark. The extension can be also focused to methods for detection of AI content in other modalities (or mixed modalities), such as images, voice or videos, which would make it even more universal.

\section{Conclusion}
\label{sec:conclusion}

The machine-generated text detection belongs to the key challenges connected with the advancements of large language models for prevention of misuse of high-quality text generation capability. The proposed IMGTB framework unifies the evaluation of the existing detection methods and simplifies comparison of new detection methods to the state-of-the-art. With a plug-and-play testing ability of new methods, research hypotheses can be easily examined. The framework can also be used for evaluation of state-of-the-art detection methods on custom data to identify the best performing one to be further used for some specific application. Automated results analysis and methods comparison also enables less proficient users to interpret the results and make a selection.
The framework reduces unnecessarily redundant work of researchers and enables them to focus their effort towards development of more effective detection methods. This can eventually accelerate the research in machine-generated text detection to catch up with the text generation, currently in the lead.

\section*{Limitations}

Besides the limitations mentioned in Section~\ref{sec:extension}, there are multiple MGTD methods using different pipelines (such as Grover by~\citealp{zellers2019neuralfakenews} or FAST by~\citealp{zhong-etal-2020-neural}), which are not yet compatible with the framework. Further work is required to integrate them. There also exist many zero-shot online services (usually paid) for MGTD, available by a custom API (application programming interface). Only one of them, GPTZero\footnote{\url{https://gptzero.me/}}, is currently supported by the framework. These were out of scope of the current work.

\section*{Ethical Considerations}

We believe that there is only a limited possibility of \textbf{misuse of our framework}. By easily identifying the most successful detection methods, the focus of malicious actors can be moved towards them in order to find ways to avoid detection. Although the mentioned risk is serious, the benefits of the provided framework mentioned in the introduction surpass such risks. The detection methods are already available, we just provide means to compare their performance.

There are additional potential ethical risks associated with the MGTD in general, such as difficulty to differentiate between malicious and legitimate use of machine-generated texts, potential harm caused by false positives or over-reliance on the results of an automated detection methods. However, these pertain more to the deployment of an MGTD service rather than to the benchmarking framework, and are therefore deemed out of scope of this work.

\section*{Acknowledgements}
This work was partially supported by the \textit{VIGILANT - Vital IntelliGence to Investigate ILlegAl DisiNformaTion}, a project funded by the European Union under the Horizon Europe, GA No. \href{https://doi.org/10.3030/101073921}{101073921}, and by \textit{vera.ai - VERification Assisted by Artificial Intelligence}, a project funded by European Union under the Horizon Europe, GA No. \href{https://doi.org/10.3030/101070093}{101070093}.

\bibliography{anthology,custom}

\end{document}